\documentclass{Interspeech2024}

\interspeechcameraready 

\title{Inconsistency-Aware Cross-Attention for Audio-Visual Fusion in \\Dimensional Emotion Recognition}

\name[affiliation={}]{R. Gnana}{Praveen}
\name[affiliation={}]{Jahangir}{Alam}

\address{
  Computer Research Institute of Montreal}
\email{ }


\begin{document}

\maketitle

\begin{abstract}
Leveraging complementary relationships across modalities has recently drawn a lot of attention in multimodal emotion recognition. Most of the existing approaches explored cross-attention to capture the complementary relationships across the modalities. 
However, the modalities may also exhibit weak complementary relationships, which may deteriorate the cross-attended features, resulting in poor multimodal feature representations. 
To address this problem, we propose Inconsistency-Aware Cross-Attention (IACA), which can adaptively select the most relevant features on-the-fly based on the strong or weak complementary relationships across audio and visual modalities. Specifically, we design a two-stage gating mechanism that can adaptively select the appropriate relevant features to deal with weak complementary relationships. 
Extensive experiments are conducted on the challenging Aff-Wild2 dataset to show the robustness of the proposed model.
\end{abstract}

\section{Introduction}
Automatic Emotion Recognition (ER) 
is a challenging problem 
due to the complex and extremely diverse nature of expressions across individuals and cultures. 
Though emotion classification
has been widely explored in the literature \cite{9039580}, 
they fail to capture the wide range of expressions on a finer granularity. Depending on the granularity of labels, regression of expressions can be formulated as 
ordinal (discrete) regression or dimensional (continuous) regression. 
Dimensional ER is even more challenging due to the high level of ambiguity in dimensional labels, often considered as valence and arousal.
Valence reflects the wide range of emotions in the dimension of pleasantness, from being negative (sad) to positive (happy). On the other hand, arousal spans a range of intensities from passive (sleepiness) to active (high excitement). 
In this paper, we have focused on the problem of 
dimensional ER using audio-visual (A-V) fusion in the valence-arousal space.

\begin{figure}[!t]
\centering
\includegraphics[width=1.0\linewidth]{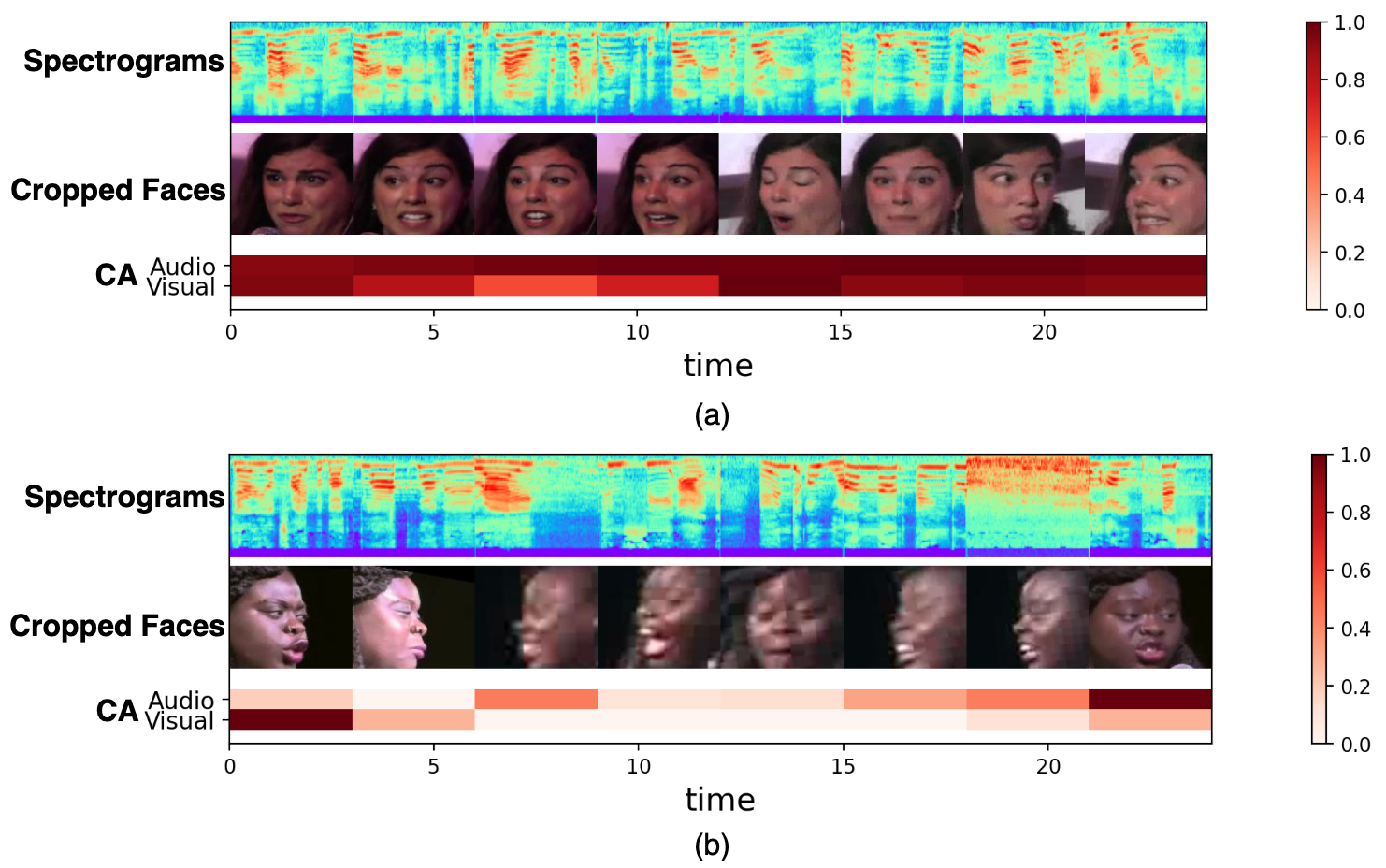}
\caption{Attention scores are normalized between 0 and 1. (a) Cross-attention scores for the subject named "12-24-1920x1080" of Affwild2 dataset. Both the modalities exhibit higher attention scores due to their strong complementary nature (portraying significant expressions). (b) Cross-attention scores of subject named "21-24-1920x1080" in Affwild2 dataset. Here, facial modality is corrupted due to extreme pose and blur, but vocal expressions are uncorrupted exhibiting rich emotions. Attending noisy face to rich vocal expressions result in lower attention scores, thereby losing the rich vocal expressions.}
\label{Demo}
\vspace{-5mm}
\end{figure}

Recently, 
multimodal ER has achieved impressive performance by leveraging complementary relationships across audio and visual modalities in videos using Cross-Attention (CA)   
\cite{praveen2023recursive,10005783,9667055,Praveen_2022_CVPR}. Though CA has been gaining momentum in capturing the complementary relationships, 
audio and visual modalities may not always pose strong complementary relationships with each other, they may also exhibit 
weak complementary relationships 
\cite{9552921}.
It has been shown that the audio and visual modalities may demonstrate conflicting (when one of the modalities is noisy or paradoxical) or dominating (when one of the modalities is restrained or unemotional) relationships with each other for ER \cite{9552921}. 
When one of the modalities is extremely noisy or restrained (weak complementary relationships), leveraging the noisy modality to attend to a good modality may deteriorate the fused 
A-V feature representations \cite{9706879}. 
For a better understanding of the problem of weak complementary relationships for ER, we have provided an interpretability analysis by visualizing the attention scores of CA as shown in Fig 
\ref{Demo}. When the modalities demonstrate strong complementary relationships, we can observe higher attention scores, thereby leveraging the complementary relationships (top image). On the other hand, when they do not strongly complement each other (bottom image), attending the rich vocal expressions with the noisy facial modality results in lower attention scores for audio, thereby degrading the audio representations also, losing the rich vocal expressions. 
Motivated by this insight, we investigate the prospect of adaptively choosing the most appropriate features based on their inter-modal relationships. 


In this work, we propose an Inconsistency-Aware Cross-Attention (IACA) model, that focuses on intra-modal relationships using Self-Attention (SA) for weak complementary scenarios, while retaining the benefits of strong complementary relationships using CA.
In particular, we propose a two-stage gating mechanism to collaboratively fuse the audio and visual features by automatically selecting the cross-attended or self-attended features based on strong or weak complementary relationships respectively. 
The proposed model can also be adapted to any variant of the CA model, adding more flexibility to the CA framework by handling the weak complementary relationships.  
To the best of our knowledge, this is the first work to investigate the inconsistency issues (weak complementary relationships) across audio and visual modalities for ER. 
The major contributions of the paper can be summarized as follows: 
(i) We investigate the potential of CA and showed that weak complementary relationships degrade the audio-visual feature representations for dimensional ER. (ii) To deal with weak complementary relationships, we propose an IACA model, a two-stage gating mechanism to adaptively select the most appropriate features on-the-fly based on the strong or weak complementary relationships. 
(iii) A detailed set of experiments are conducted to evaluate the proposed model on the challenging Aff-Wild2 dataset. The proposed model achieves consistent improvement over multiple variants of CA models, achieving state-of-the-art performance for dimensional ER. 

\section{Related Work}



\label{sec:format}
Most conventional approaches for dimensional ER explored simple feature concatenation \cite{9320301,7477679} or LSTM-based fusion models \cite{cite7,cite6}. In video-based ER, leveraging complementary relationships across the audio and visual modalities was found to be promising in outperforming the unimodal systems \cite{LIU2023679}. 
With the massive success of transformers, several approaches have been proposed to exploit CA with transformers and found significant improvement for ER \cite{9928357}. Parthasarathy et al. \cite{srini_2021_SLT} explored multimodal transformers, where CA is integrated with the SA module to obtain the A-V cross-modal feature representations. 
Praveen et al. \cite{9667055} proposed a CA model to effectively capture the complementary relationships by allowing the modalities to interact with each other. They have further extended the approach by introducing joint feature representation to the CA framework 
\cite{10005783, Praveen_2022_CVPR} and recursive fusion \cite{praveen2023recursive}. Although these methods 
have shown remarkable success with CA, they rely on the assumption that the audio and visual modalities always strongly complement each other. When they pose weak complementary relationships,
these methods will result in poor performance.

Gating mechanisms with attention have been widely explored for multimodal fusion to control the flow of modalities based on their importance 
to reduce redundancy \cite{10.1007/s00521-019-04559-1,10.1145/3382507.3418867} or noisy modalities \cite{9053012,jiang2022crossmodality}.
Aspandi et al. \cite{9726856} proposed a gated-sequence neural network, 
where the gating mechanism is explored for temporal modeling 
to adaptively fuse the modalities based on their relative importance. 
Kumar et al. \cite{9053012} explored the conditional gating mechanism using a nonlinear transformation by modulating the cross-modal interactions to learn the relative importance of modalities. 
Jiang et al. \cite{jiang2022crossmodality} also explored CA with a forget gate to discard the redundant information and focus on salient cross-modal representations for multimodal sentiment analysis. 
Unlike prior approaches, we have explored a conditional two-stage gating mechanism to tackle the problem of weak complementary relationships by dynamically selecting the semantic features pertinent to 
the synergic A-V relationships. 

\begin{figure*}[!t]
\centering
\includegraphics[width=1.0\linewidth]{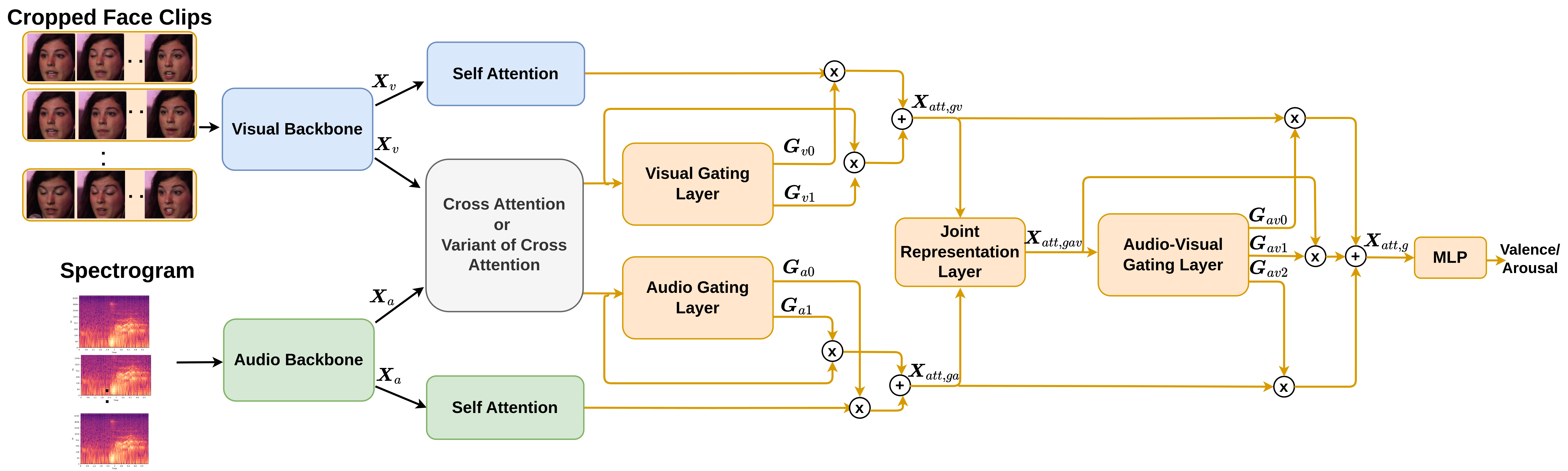}
\caption{Illustration of the proposed Inconsistency-Aware Cross-Attention model.}
\label{AdapJA}
\end{figure*}

\section{Proposed Approach}
\label{sec:pagestyle}
\noindent \textbf{A) Notations:}
For an input video sub-sequence $S$, $L$ non-overlapping video clips are uniformly sampled, and the corresponding deep feature vectors are obtained from the pre-trained models of audio and visual modalities. Let ${\boldsymbol X}_{a}$ and ${\boldsymbol X}_{v}$ denote the deep feature vectors of audio and visual modalities respectively for the given input video sub-sequence $\boldsymbol S$ of fixed size, which is expressed as 
${ \boldsymbol X}_{a}=  \{  x_{a}^1,  x_{a}^2, ...,  x_{ a}^L \} \in \mathbb{R}^{d\times L}$ 
and 
${ \boldsymbol X}_{\mathbf v}=  \{  x_{ v}^1,  x_{ v}^2, ...,  x_{ v}^L \} \in \mathbb{R}^{d\times L}$
where $d$ represents the dimension of the audio and visual feature vectors, and $ x_{ a}^{ l}$ and $ x_{ v}^{ l}$ denote the audio and visual feature vectors of the video clips, respectively, for $l = 1, 2, ..., L$ clips.
\noindent \textbf{B) Preliminary-Cross Attention:}
In this section, we briefly introduce the CA \cite{9667055} (baseline fusion model) as a preliminary to the proposed IACA model.  
Given the audio and visual feature vectors ${\boldsymbol X}_{ a}$ and ${\boldsymbol X}_{ v}$ for a video sub-sequence $\boldsymbol S$, the cross-correlation across the modalities are computed as 
$\boldsymbol Z={\boldsymbol X}_{ a}^\top \boldsymbol W{\boldsymbol X}_{ v}$ ($\boldsymbol W$ denotes cross-correlation weights).
Next, we compute the CA weights of audio and visual features, ${\boldsymbol A}_{ a}$ and ${\boldsymbol A}_{ v}$ by applying column-wise softmax of $\boldsymbol Z$ and $\boldsymbol Z^\top$, respectively: 
\begin{align}
{\boldsymbol A}_{{ a}_{i,j}}
=\frac{ e^{{\mathbf Z}_{i,j}}}{\overset{ K}{\underset{ k\boldsymbol=1}{\sum}} e^{{\mathbf Z}_{k,j}}}    
\quad  \text{and} \quad
{\boldsymbol A}_{{ v}_{i,j}}
=\frac{ e^{{\mathbf Z^\top}_{i,j}}}{\overset{K}{\underset{ k\boldsymbol=1}{\sum}} e^{{\mathbf Z^\top}_{i,k}}}    
\end{align}


After obtaining the CA weights, they are used to compute the attention maps of the audio and visual features to make them more comprehensive and discriminative as: 
\begin{align}
\widehat{\mathbf{X}_{a}}=\mathbf{X}_{a}{{\mathbf A}_{ a}} \quad  \text{and}  \quad
\widehat{\mathbf{X}_v}=\mathbf X_v{{\mathbf A}_{ v}}
\end{align}
The attention maps are then added to the corresponding features to obtain the attended features, which is given by:  
\begin{align}
{\boldsymbol X}_{att,a}= \tanh({\boldsymbol X}_{ a}+\widehat{\mathbf X_a}) \\
{\boldsymbol X}_{att,v}= \tanh({\boldsymbol X}_ v+\widehat{\mathbf X_v})
\end{align}
\noindent \textbf{C) Inconsistency-Aware Cross-Attention Model:}
To deal with weak complementary relationships, we design two-stage gating layers, where the first stage selects the cross-attended or self-attended features based on strong or weak complementary relationships, respectively for the individual modalities. In the second stage, we choose the most relevant features among the joint A-V features and the features of individual modalities to mitigate the impact of extremely noisy modality. The gating layers of the first stage are designed using a Fully Connected (FC) layer with two output units, one for the cross-attended feature and one for the self-attended feature, which is given by  
\begin{align}
{\boldsymbol W}_{go,v} = {\boldsymbol X}_{ att,v}^\top \boldsymbol W_{gl,v}
\hspace{-2mm} \quad \text{and} \hspace{-2mm} \quad {\boldsymbol W}_{go,a} = {\boldsymbol X}_{att,a}^\top \boldsymbol W_{gl,a}
\end{align}
where $\boldsymbol W_{gl,a},\boldsymbol W_{gl,v} \in\mathbb{R}^{d\times 2}$, 
are the learnable weights of the gating layers and $\boldsymbol W_{go,a} \in\mathbb{R}^{L\times 2}$, $\boldsymbol W_{go,v} \in\mathbb{R}^{L\times 2}$ are outputs of gating layers of audio and visual modalities, respectively. To obtain probabilistic attention scores, the output of the gating layers is activated using a softmax function with a small temperature $T$, as given by 
\begin{align}
{\boldsymbol G}_{{a}}
=\frac{ e^{{\boldsymbol W}_{go,a}/T}}{\overset{ K}{\underset{ k\boldsymbol=1}{\sum}} e^{{\boldsymbol W}_{go,a}/T}}    
\quad  \text{and} \quad
{\boldsymbol G}_{{v}}
=\frac{ e^{{\boldsymbol W}_{go,v}/T}}{\overset{K}{\underset{ k\boldsymbol=1}{\sum}} e^{{\boldsymbol W}_{go,v}/T}}    
\end{align}
where $\boldsymbol {G}_{a} \in\mathbb{R}^{L\times 2}$, $\boldsymbol {G}_{v}\in\mathbb{R}^{L\times 2}$ denotes the probabilistic attention scores of audio and visual modalities, respectively. $K$ denotes the number of output units of the gating layer (here $K=2$). 
The probabilistic attention scores of the gating layer help to estimate the relevance of cross-attended or self-attended features. 
The two columns of $\boldsymbol G_a$ correspond to the 
scores of self-attended features (first column) and cross-attended features (second column). To multiply with the corresponding feature vectors, each column is replicated to match the dimension of the corresponding feature vectors, 
denoted by $\boldsymbol G_{a0}$, $\boldsymbol G_{a1}$ and $\boldsymbol G_{v0}$, $\boldsymbol G_{v1}$ for audio and visual modalities, respectively, 
which is further fed to the ReLU activation function as: 
\begin{equation}
{\boldsymbol X}_{att,gv} = \text{ReLU}(\boldsymbol X_{v} \otimes \boldsymbol G_{v0} + {\boldsymbol X}_{att,v} \otimes \boldsymbol G_{v1})  
\end{equation}   
\begin{equation}
{\boldsymbol X}_{att,ga} = \text{ReLU}(\boldsymbol X_{a} \otimes \boldsymbol G_{a0} + {\boldsymbol X}_{att,a} \otimes \boldsymbol G_{a1})  
\end{equation}  
where $\otimes$ denotes element-wise multiplication. Now the attended features from the gating layers ${\boldsymbol X}_{att,ga}$ and ${\boldsymbol X}_{att,gv}$ are 
fed to the joint representation layer (concatenation + FC) 
to match the dimension $d$ of individual modalities, 
denoted as 
${\boldsymbol X}_{att,gav}$.  
In the second stage, we compute the attention scores for the features ${\boldsymbol X}_{att,ga}$, ${\boldsymbol X}_{att,gv}$ and ${\boldsymbol X}_{att,gav}$ using A-V gating layer, where the gating layer is designed with 3 output units. 
The A-V gating layer helps to mitigate the adverse effects of severely corrupted or missing modalities and focus more on the uncorrupted or contributing modality. Similar to the first gating layer, the softmax function is used and the probabilistic attention scores are replicated to multiply with the corresponding feature vectors, which is given by 
\begin{equation}
\begin{aligned}
{\boldsymbol X}_{att,g} = \text{ReLU}(\boldsymbol X_{att,ga} \otimes \boldsymbol G_{av0} + {\boldsymbol X}_{att,gv} \otimes \boldsymbol G_{av1}  \\ \hspace{-7mm} + {\boldsymbol X}_{att,gav} \otimes \boldsymbol G_{av2})  
\end{aligned}
\end{equation}
where $\boldsymbol G_{av0}$, $\boldsymbol G_{av1}$ and $\boldsymbol G_{av2}$ denote the attention scores of ${\boldsymbol X}_{att,ga}$, ${\boldsymbol X}_{att,gv}$ and ${\boldsymbol X}_{att,gav}$ respectively. Finally, the feature vector $\boldsymbol X_{att,g}$ obtained from the A-V gating layer is
fed to the 
Multi Layer Perceptron (MLP) to obtain the final prediction of valence or arousal. 

\section{Results and Discussion}
\label{sec:typestyle}
\noindent \textbf{A) Datasets:}
Aff-Wild2 is the largest dataset, consisting of $594$ videos collected from YouTube, all captured in the wild \cite{10208745}. The annotations are provided by four experts using a joystick and the final annotations are obtained as the average of the four raters. 
Sixteen of these videos display two subjects, both of which have been annotated. In total, there are $2,993,081$ frames with $584$ subjects and 
annotations for valence and arousal are provided continuously in the range of $\lbrack-1,1\rbrack$. The dataset is divided into 360, 72, and 162 videos in training, validation, and test sets, respectively in a subject-independent manner. 

\noindent \textbf{B) Implementation Details:}
The audio and visual modalities are preprocessed and the feature vectors are obtained from pretrained backbones similar to that of the baselines \cite{9667055,10005783,praveen2023recursive}. Following the baselines, we have used R3D \cite{8578773} network pre-trained on the Kinetics-400 dataset for visual modality and Resnet-18 \cite{7780459} for audio modality. 
The implementation details such as hyperparameters, optimizer, sequence lengths, etc for the individual modalities, including the loss function of the proposed framework along with the baselines are retained from the source codes provided by the authors \cite{9667055,10005783,praveen2023recursive} with their default setting. 
Concordance Correlation Coefficient (CCC) is the widely used evaluation metric in the literature of dimensional ER to measure the level of agreement between the predictions and ground truth of valence and arousal \cite{cite7,9857097,praveen2023recursive,10005783}. 
\begin{table*}[!h]
\centering
  \caption{CCC performance of the proposed model in comparison to the state-of-the-art methods}. 
  
  \label{tab:comparisiontoSOAonAffwild2}
  \begin{tabular}{|l|c|c|c|c|c|c|c}
   \hline
 \textbf{Method}& 
 \textbf{Type of } & \multicolumn{2}{|c|}{\textbf{Validation Set}} & \multicolumn{2}{|c|}{\textbf{Test Set}} \\
    \cline{3-6}
     & \textbf{Fusion} & \textbf{Valence} & \textbf{Arousal} & \textbf{Valence} & \textbf{Arousal} \\
\hline \hline
    Zhang et al. \cite{9607460} & Leader Follower Attention & 0.469 &  0.649 & 0.463 & 0.492  \\ \hline
    Karas et al. \cite{Karas_2022_CVPR} & LSTM + Transformers & 0.388 & 0.551 & 0.418 & 0.407 \\ \hline
    Meng et al. \cite{9857097} & LSTM + Transformers & 0.605 & 0.668 & 0.606 & 0.596 \\ \hline
     Zhou et al. \cite{Zhou_2023_CVPR} & TCN + Transformers &0.550 & 0.680 & 0.566 & 0.500\\ \hline
    Zhang et al. \cite{Zhang_2023_CVPR} &  Transformers &0.648 & 0.705 & 0.523 & 0.545\\ \hline
    Praveen et al. \cite{10005783}& JCA & 0.657  & 0.580 & 0.451 & 0.389\\ \hline
    Praveen et al. \cite{praveen2023recursive} &
    RJCA & 0.721 & 0.694 & 0.467 & 0.405\\ \hline
    Ours & RJCA + IACA & \textbf{0.749} &  \textbf{0.725} & \textbf{0.505} & \textbf{0.474}\\ \hline
  \end{tabular}
\end{table*}

\noindent \textbf{C) Ablation Study:}
In order to analyze the impact of the proposed IACA model on different variants of CA, we have considered
four baseline models: CA \cite{9667055}, Joint Cross-Attention (JCA) \cite{10005783}, Recursive Joint Cross-Attention (RJCA) \cite{praveen2023recursive}, and Transformer-based Cross-Attention (TCA) \cite{srini_2021_SLT}. 
The proposed IACA model is added to each of the baselines to analyze the performance improvement. We can observe that the proposed model shows consistent improvement over all the baselines. Of all these baselines, CA and TCA rely only on cross-modal interactions, whereas JCA and RJCA exploit both intra- and inter-modal relationships using joint feature representation in the CA framework. Since the proposed model is focused on dealing with weak complementary inter-modal relationships, there is a significant performance improvement (relative) for CA and TCA than that of JCA and RJCA as shown in Table \ref{Ablation Study}. 
\begin{figure}[!t]
\hspace{-10mm}
\includegraphics[width=1.2\linewidth]{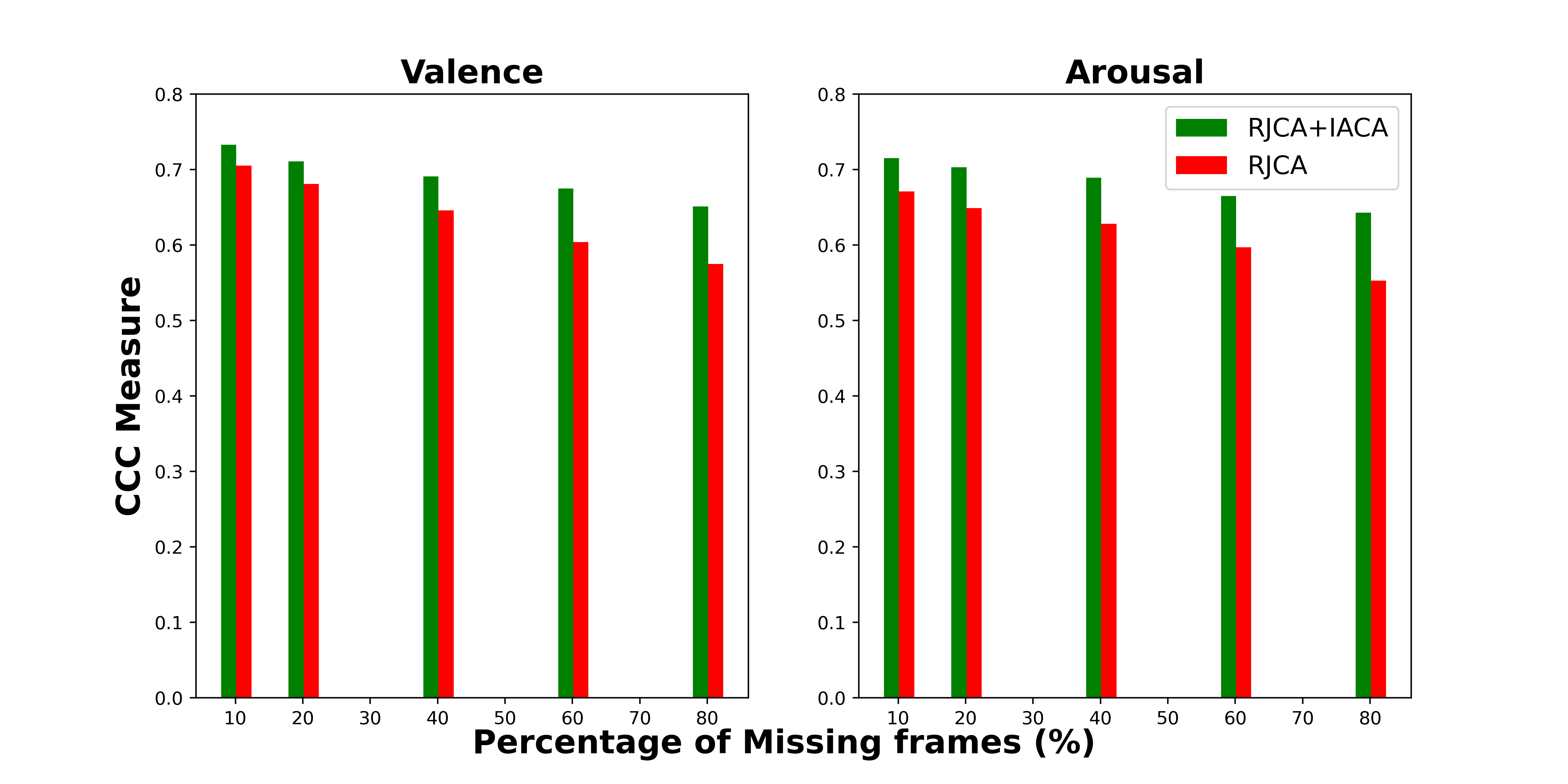}
\caption{CCC Performance of the proposed model along with RJCA with a growing proportion of missing audio modality}
\label{AdapJA}
\end{figure}

The proposed model has also been evaluated under the scenario of missing modality, in which a growing portion of audio modality is replaced by "zeros" to replicate the scenario of missing modality. Specifically, we chose the video sample "131-30-1920x1080" with 8012 frames and the audio modality (spectrograms) is dropped 
at varying percentages of 10, 20, 40, 60, and 80 in test mode. The CCC performance of the proposed approach is compared to that of RJCA for valence and arousal, where RJCA is considered as the baseline. 
We can observe that the performance of the proposed model has not decreased significantly compared to the baseline (RJCA) for both valence and arousal as depicted in Fig \ref{AdapJA}. By exploiting the two-stage gating mechanism, the proposed model effectively handles the scenario of missing modality. On the contrary, the performance of the RJCA model declines significantly beyond 50\% of missing audio frames.

\begin{table}[h]
\small
\centering
\caption{CCC performance of the proposed model with multiple variants of CA model on validation set. $\Delta$ represents relative improvement in percentage. w/ \& w/o stand for with and without, respectively.}
\label{Ablation Study}
\begin{tabular}{|l|c|c|c|c||c|c|c|c|c|c|} 
\hline
\textbf{Method}  & \textbf{Valence} &  \textbf{Arousal} \\
\hline \hline
CA w/o IACA  & 0.541 & 0.517  \\
\hline
~CA w/ IACA & 0.632 & 0.597 \\
\hline
~~~~~~~~~~$\Delta$ & \textbf{16.8\%} & \textbf{15.5\%} \\
\hline
\hline
TCA w/o IACA & 0.564 &  0.543 \\
\hline	
~TCA w/ IACA & 0.637 & 0.629  \\
\hline
~~~~~~~~~~$\Delta$ & \textbf{12.9\%}  & \textbf{15.8\%} \\
\hline \hline
JCA w/o IACA & 0.657 & 0.580  \\
\hline
JCA w/ IACA & 0.693 & 0.609  \\
\hline
~~~~~~~~~~$\Delta$ & \textbf{5.5\%} & \textbf{5.0\%} \\
\hline \hline
RJCA w/o IACA & 0.721 & 0.694  \\
\hline
RJCA w/ IACA & 0.749 & 0.725 \\
\hline
~~~~~~~~~~$\Delta$ & \textbf{3.9\%} & \textbf{4.5\%} \\
\hline 
\end{tabular}
\end{table}

\noindent \textbf{D) Comparison to state-of-the-art:}
Since most of the relevant approaches on this dataset have been submitted to the ABAW challenges \cite{kollias2021analysing,kollias2022abaw}, we compared the proposed approach with the relevant state-of-the-art models for A-V fusion in dimensional ER. Zhang et al. \cite{9607460} explored a leader-follower attention strategy by exploiting the audio modality to attend to the visual modality to boost the performance of the visual modality. However, they do not show major improvement in fusion performance over the uni-modal performances. Karas et al. \cite{Karas_2022_CVPR} and Meng et al. \cite{9857097} explored the ensembling of fusion models with LSTMs and transformers, where \cite{9857097} significantly improved the performance on both validation and test sets by leveraging multiple backbones for each modality and three external datasets to achieve better generalization. Zhou et al. \cite{Zhou_2023_CVPR} and Zhang et al. \cite{Zhang_2023_CVPR} also exploited transformer models, where the feature vectors of audio and visual modalities are concatenated and fed to the transformer model. \cite{Zhou_2023_CVPR} used Temporal Convolutional Networks along with transforms, whereas \cite{Zhang_2023_CVPR} leveraged large-scale pretraining using Masked Auto-encoders. Most of these methods \cite{9857097,Zhou_2023_CVPR,Zhang_2023_CVPR} do not effectively capture the inter-modal relationships across the modalities and are geared towards achieving higher performance on the test set by exploiting large-scale pretraining on additional datasets and multiple backbones for each modality.

Unlike these methods, \cite{10005783} and \cite{praveen2023recursive} focused on improving the fusion performance by effectively capturing the inter-modal relationships, still retaining the intra-modal relationships. 
Both \cite{10005783} and \cite{praveen2023recursive} improved the fusion performance by introducing the joint feature representation in the CA framework to capture both intra- and inter-modal relationships. Though they did not achieve significant improvement on the test set as they used only single backbones for each modality without any external datasets, their performance is solely attributed to the sophisticated fusion model 
by effectively leveraging the intra- and inter-modal relationships. 
Since the primary focus of our work is to improve fusion performance by handling weak complementary relationships and severely corrupted modalities, the proposed model has been validated on these sophisticated CA-based fusion models \cite{praveen2023recursive}. We can observe that the fusion performance of RJCA \cite{praveen2023recursive} has been further improved on both validation and test sets 
using the proposed IACA model 
as shown in Table \ref{tab:comparisiontoSOAonAffwild2}.


\section{Conclusion}
In this paper, we investigated the prospect of retaining the benefits of strong complementary relationships across audio and visual modalities, while being robust to weak complementary relationships. 
In particular, we proposed an IACA model using a two-stage gating mechanism, where the first stage deals with weak complementary relationships by selecting the cross-attended or self-attended features based on strong or weak complementary relationships, respectively. The features obtained from the first gating layer is further refined by focusing on the significant modality to suppress the impact of extremely corrupted or missing modalities. 
Extensive experiments were conducted to analyze the impact of the proposed IACA model on multiple variants of Cross-Attention (CA) and found consistent improvement on both valence and arousal. Leveraging sophisticated backbones for audio and visual modality is expected to further improve the performance of the model on the test set. 





\bibliographystyle{IEEEtran}
\bibliography{mybib}

\end{document}